\documentclass{article} 
\usepackage[preprint]{colm2026_conference}

\usepackage{microtype}
\usepackage{hyperref}
\usepackage{url}
\usepackage{booktabs}
\usepackage{microtype}
\usepackage{hyperref}
\usepackage{url}
\usepackage{booktabs}
\usepackage{graphicx}
\usepackage{amsmath}
\usepackage{amsmath}
\DeclareMathOperator*{\argmax}{arg\,max} 
\usepackage{booktabs}
\usepackage{longtable}
\usepackage{graphicx}
\usepackage{subcaption}
\usepackage{float}

\definecolor{darkblue}{rgb}{0, 0, 0.5}
\hypersetup{
  colorlinks=true,
  citecolor=darkblue,
  linkcolor=darkblue,
  urlcolor=darkblue
}

\usepackage{lineno}

\definecolor{darkblue}{rgb}{0, 0, 0.5}
\hypersetup{colorlinks=true, citecolor=darkblue, linkcolor=darkblue, urlcolor=darkblue}

\title{Readable but Not Controllable: Neuron-Level Evidence for Medical LLM Hallucination}

\author{Vijay Vankadaru$^1$, Asha Matthews$^1$, Tanya Roosta$^{*, 1}$ \& Peyman Passban \thanks{This work was conducted independently and is not associated with the author’s employer. All views and
conclusions are solely those of the author.} \\
$^1$ School of Information\\
University of California, Berkeley\\
California, USA \\
\texttt{vankadaruvijay@ischool.berkeley.edu, passban.peyman@gmail.com}
}

\begin{document}

\ifcolmsubmission
\linenumbers
\fi

\maketitle

\begin{abstract}
Hallucination remains one of the central obstacles to deploying medical LLMs. Yet, even when hallucination can be detected, it is still unclear whether the internal representations associated with it can be used for control rather than detection alone. Using four open-source models across a suite of medical question-answering datasets, we show that a simple, carefully conditioned probe can reliably detect hallucination, with AUROC scores between 0.77 and 0.86 in our case. We further show that this signal is distributed and redundant rather than narrowly localized. Systematically selected neurons outperform random neurons only at very small subset sizes, whereas random subsets of a few hundred neurons recover nearly the full signal, and low-dimensional random projections preserve most of the detection performance. Beyond detection, we test whether this representation is causally actionable. Across 16 model--dataset combinations, our results reveal a sharp gap between decodability and controllability. The same internal structure that makes hallucination easy to detect does not translate into reliable neuron-level control. These findings show that medical hallucination seems to be readily visible in internal activations, but not easily corrected by steering the neurons most associated with it. More broadly, our results suggest that hallucination mitigation is not simply a matter of identifying the right neurons, and point to a deeper separation between what representations reveal and what they allow us to change.
\end{abstract}

\section{Introduction}\label{intro}
This paper studies whether factual hallucination in medical LLMs can be understood through the lens of hallucination-associated neurons. Specifically, we investigate a mechanism for detecting neurons that may help differentiate truthful answers from hallucinated ones. Once these neurons are selected, we can use them to classify internal patterns and behaviors of LLMs and, hopefully, intervene in their hallucinations. For this purpose, we simply measure the contribution of each neuron to the output of its associated layer. 

A common technique to measure this type of contribution is CETT \citep{zhang2024relu}, which is (almost) equivalent to the magnitude of a neuron and shows how much a neuron contributes to the model’s generated tokens. The closest paper that relies on a CETT-style analysis is H-Neuron \citep{gao2025h}. In that work, the authors identify neurons whose contribution patterns are strongly linked to hallucinated answers. In this respect, our work follows a similar direction. However, we ask a more specific question. \textit{Are hallucination-associated activations in medical models merely detectable, or can they also serve as causal control levers?} In doing so, we move beyond H-Neuron and investigate the possibility of causal control.

H-Neuron suggests that some behaviors, such as over-compliance, are linked to a small set of important neurons and can show causal effects when those neurons are changed. In contrast, this project studies factual hallucination in medical LLMs and finds that the signal is spread across hundreds of neurons. The signal can be detected, but changing those neurons does not necessarily change the model’s hallucination behavior. Our experimental results attempt to dissect why this happens. Throughout the paper, further differences among the problems we investigate and others will become clearer as we provide more technical details.

One of the key findings of our research is that hallucination is clearly detectable from internal activations, but this detectability does not translate into reliable causal control. This suggests that fixing hallucination is not a straightforward, task-based problem. This idea is comprehensively discussed by \citet{xu2024hallucination} and \citet{matthewsrethinking}. The main interpretation is that hallucination in these models is legible, which is useful because it can at least inform the user, but it is not yet actionable. We hope that research like ours can shed light on this problem and help the field move toward solutions faster.

Our contributions are threefold. \textit{\textcolor{blue}{i}}) We provide a mechanism-aware, neuron-based study of hallucination in medical-specialized LLMs. \textit{\textcolor{blue}{ii}}) We show that hallucination is represented as a distributed and low-dimensional signal. The full k-sweep (see next sections for details), shows that random subsets of a few hundred neurons can match selected subsets and can outperform the full layer. \textit{\textcolor{blue}{iii}}) We also demonstrate a clear detection-causation decoupling. A hallucination signal can be highly detectable to a linear probe while failing to act as a reliable control lever under activation steering.

\section{Background}\label{lit_review}
A growing body of research suggests that hallucination- and truthfulness-related behavior can be detected from the internal activations of LLMs. SAPLMA \citep{azaria2023internal} shows that hidden-state probes can predict whether model-generated statements are truthful. Unsupervised methods such as contrast-consistent search (CCS) \citep{burns2022discovering} and HaloScope \citep{du2024haloscope} suggest that hallucination signals can be recovered from activation space without relying only on manually labeled examples. Other methods, including semantic entropy, SelfCheckGPT \citep{manakul2023selfcheckgpt}, and mass-mean probing, estimate uncertainty, factuality, or hallucination risks from outputs and/or internal states.

Our work focuses on hallucination detection in medical question answering (Q\&A) and asks whether features that support detection also provide effective targets for intervention. As shared in the previous section, the work most directly related to ours is H-Neurons, which uses neuron ranking and sparse selection to identify hallucination-associated neurons. It reports that fewer than 0.1\% of feed-forward neurons can predict hallucination behavior, however, its strongest causal result concerns over-compliance rather than factual hallucination. Recent cross-domain work by \citet{vaddi2026hallucination} further shows that selected hallucination neurons transfer poorly across domains, which indicates that these signals may be distributed, redundant, or domain-dependent. This is also what we observed. 

This distinction is important because probes (also known as hallucination detection models) can reveal that information is encoded in a representation, but not whether the model actually uses that information to generate its output. Mechanistic interpretability work emphasizes that causal claims require interventions such as activation patching or representation steering. Related control methods, including inference-time intervention \citep{li2023inference} and contrastive activation addition \citep{panickssery2023steering}, show that modifying activation directions can steer high-level behavior. These methods motivate our experiments, but our results demonstrate that hallucination-associated neurons selected for detection do not reliably provide effective control targets. This may be due to the nature of our scope.

Medical hallucination is especially important because fluent but incorrect outputs can mislead users in high-stakes healthcare contexts. We therefore evaluate medical Q\&A-style datasets to test whether internal hallucination signals are stable across settings. External mitigation methods, including retrieval-augmented generation \citep{ayala2024reducing}, GraphCheck \citep{chen2025graphcheck}, and MedReflect \citep{huang2025medreflect} improve factual grounding through evidence retrieval, structured fact checking, and self-reflection, respectively. These approaches are complementary to internal-activation methods, but they do not explain where hallucinations arise inside the model or whether they can be directly controlled.

Overall, prior work shows that hallucination-related information is often present in model activations and that activation-based methods can sometimes steer model behavior. However, the relationship between detection and control remains underexplored, especially in medical domains. We show that medical hallucinations are reliably decodable, but the signal is distributed and neuron-level targets do not meaningfully outperform random neuron interventions. These findings support the view that features useful for detection are not necessarily sufficient for controlling or perhaps the intervention should be addressed at a level higher than individual neurons. 

\section{Methodology}\label{method}

\subsection{Detection Mechanism}\label{detection}
We compare activations from correct and hallucinated generations, then compute neuron contribution to measure which neurons matter most. These neuron-level features are fed into a probe (a simple classifier), to detect which neurons are most predictive of hallucination. To calculate contribution, we tested two methods. We use $\texttt{raw}$ values, which are the direct post-activation values of the neurons, specifically the SwiGLU post-activations ($\texttt{act\_fn}$). Therefore, in our setup $\texttt{raw} = \texttt{act\_fn}(\texttt{gate\_proj}(x))$, where $\texttt{gate\_proj}$ provides the gating information inside the MLP/feed-forward block and $x$ refers to the input. We also use $\texttt{exact\_cett}$, the true neuron contribution to the residual stream, computed as:
\[
\mathrm{exact\_cett}_{j}
=
\frac{
\left|
\mathrm{\texttt{act\_fn}}(\mathrm{\texttt{gate\_proj}}(x))_{j}
\cdot
\mathrm{\texttt{up\_proj}}(x)_{j}
\right|
\,
\left\| W_{\mathrm{down}}[:,j] \right\|_2
}{
\left\| \mathrm{MLP}(x) \right\|_2
}
\]
where $j$ indexes an individual MLP neuron, $W_{\mathrm{down}}[:,j]$ is the column of the 
down-projection matrix corresponding to neuron $j$, and 
$\left\| W_{\mathrm{down}}[:,j] \right\|_2$ measures how strongly that neuron can affect the residual stream, and $\left\| \mathrm{MLP}(x) \right\|_2$ is the norm of the full MLP output, which normalizes the contribution so that neurons can be compared within the 
same layer. Our analyses show that $\texttt{raw}$ and $\texttt{exact\_cett}$ are within $0.003$ AUROC of each other almost everywhere, which means that $\texttt{exact\_cett}$ serves primarily as a comparability and normalization choice rather than providing a meaningful detection gain. 

To make sure we have enough signal to decide which neurons contribute to the process, in our setup, we experimented with $\texttt{raw}$ values, extracted $\texttt{cett}$ activations, selected $\texttt{top-k}$ neurons, $\texttt{random-k}$ neurons, and probed with $\texttt{sparse L1}$ selection. For top neuron selection, we rank neurons by their individual discriminative AUROC, computed via the Mann-Whitney rank identity:
\[
\mathrm{score}_j
=
\max\left(
\frac{
R_{1,j} - \frac{n_1(n_1+1)}{2}
}{
n_1 n_0
},
\;
1 -
\frac{
R_{1,j} - \frac{n_1(n_1+1)}{2}
}{
n_1 n_0
}
\right)
\]
where \(j\) indexes a neuron, \(R_{1,j}\) is the sum of ranks for hallucinated examples using neuron \(j\)'s activation values, \(n_1\) is the number of hallucinated examples, and \(n_0\) is the number of correct examples. The folded score \(\max(\mathrm{AUC}_j, 1-\mathrm{AUC}_j)\) treats neurons as useful whether they are higher for hallucinations or lower for hallucinations. $\texttt{top-k}$ in this setup is the k neurons with the highest individual separation between hallucinated and correct generations. As a complementary, supervised form of selection, we also fit an L1-regularized logistic regression (\texttt{liblinear} solver) over all neurons. The L1 penalty drives most coefficients to exactly zero, so the probe selects its own sparse subset, and the number of neurons it retains (its nonzero coefficients) provides a data-driven estimate of how many neurons are actually needed. This mirrors the sparse-selection strategy used to identify hallucination-associated neurons in \citet{gao2025h}.

Once these neuron signals are extracted, we look inside the model at different layers and ask whether a simple classifier can tell from the model’s internal activations whether the answer is hallucinated. To implement, a linear probe is trained on the post-activations at the best layer. The best layer is the one that is most recognizable for detecting hallucination. In other words, when the best layer is connected to the probe, it provides the highest detection accuracy. For the best model, we look at its $\texttt{Gen0}$ signal, meaning we analyze it right after the prompt is consumed. 

The probe is intentionally simple. It is a plain L2-regularized logistic regression (not a neural network). It is linear, so the AUROC measures the linear decodability of the hidden state. Its input is the per-token MLP activation vector at one layer, specifically, as mentioned, the SwiGLU post-activation at the last prompt token (the $\texttt{Gen0}$ position). Features are standardized with StandardScaler, using zero mean and unit variance, before fitting. Its output is a single scalar, $P(hallucinated)$, for one binary class (hallucinated versus correct). The probe itself has no hidden layers. It is a single linear layer with a Sigmoid. We fit it independently at every transformer layer and report the best layer. The selected layers are the per-model $\argmax$ over that layer scan. Table \ref{tab:best-gen0-layer} shows the best layers found by our probe for each model. For details about models see Section \ref{setup} as well as the appendix.
\begin{table}[h]
\centering
\begin{tabular}{lllll}
\toprule
\textbf{Model} & \textbf{BioMistral-7B} & \textbf{Meditron-7B} & \textbf{PMC-LLaMA-7B} & \textbf{Llama-3.1-8B-Instruct} \\
\midrule
\textbf{Best Layer} & L17 & L14 & L10 & L15 \\
\bottomrule
\end{tabular}
\caption{\label{tab:best-gen0-layer}Best Gen0 detection layer for each model.}
\end{table}

Using a simple probe matters because it shows that hallucination-related information is linearly accessible in the activation space and it is not the effect of a highly complex classifier. The first stage of our analysis asks whether hallucinations can be detected from internal activations.

\subsection{Geometry Analysis}\label{gm_analysis}
In the second stage of our analysis, we examine where the hallucination signal resides inside the model and how it is structured, with the goal of understanding the geometry of the phenomenon. We compare $\texttt{top-k}$ neurons with $\texttt{random-k}$ neurons to see whether the most important-looking neurons are truly special or whether random groups work almost as well. We also check whether the same neurons are selected at different token positions using overlap analysis. If the overlap is low, it means the exact neuron identities change across positions. We then use centered kernel alignment (CKA) to examine a related but broader question, namely \textit{whether the overall activation pattern remains similar even when the specific neurons involved change}. High CKA means the model may carry the same hallucination information in different neuron coordinates. 

Furthermore, we use principal component analysis (PCA), to see whether the signal is spread across many independent patterns or is mostly captured by one or a few main directions, such as PC1. We also test cross-format transfer to understand whether a detector trained in one prompt or answer format still works in another format. Overall, this geometry analysis is meant to show whether hallucination is stored in a few special neurons or instead appears as a broader, repeated, low-dimensional pattern spread across many neurons and formats.

The main finding from our analyses is that the signal is not limited to a tiny set of neurons. Therefore, we should not expect a small set of privileged neurons to detect hallucination alone. Many neurons also carry similar or overlapping information. Accordingly, removing one neuron may not influence the signal because other neurons also contain related evidence, and these neurons mostly share the same few general signals. To measure this distribution effect, we relied on a k-sweep analysis, repeating the experiment across a range of k values. Here, k is the number of neurons used by the probe. 

The k-sweep results show that selected neurons outperform random neurons mainly when k is small, especially below $\sim$50 neurons. As k increases, random subsets quickly catch up. By k $\approx$ 300, selected and random subsets converge. This means no single small set appears to be uniquely responsible for the readable signal. This can also be viewed from a redundancy perspective that signals are repeated across many neurons. Even if we randomly choose a few hundred neurons, that group often still contains enough information to detect hallucinations. Refer to experimental results in the next sections and the appendix for details. 

\subsection{Intervention Method}\label{int}
Once hallucination and its properties have been detected, we then examine if it can be controlled through intervention. Specifically, we ask whether the detected signals can be used to reduce hallucination. Our results reveal that they cannot, at least within the activation-steering family tested here. The study applies activation steering on a held-out 20\% test split. Directions are derived on the train split and evaluated on held-out examples. The study tests three ways of choosing where to change the model’s internal activations. First, it uses the neurons with the highest CETT scores. Second, it uses the PC1 direction, that seems to be the strongest overall pattern separating hallucinated and correct answers. Third, it uses the difference-of-means direction, which is found by comparing the average activation pattern for hallucinated answers with the average pattern for correct answers. We compare these methods against several controls, including random neurons, unrelated directions, and a null setting in which the intervention strength is set to zero, so that no actual change is made. These controls help determine whether the intervention is genuinely using hallucination-related information or if the observed effects emerged by chance. 

\section{Experimental Setup}\label{setup}
As mentioned previously, the main metric used to report our results is AUROC.\footnote{The Area Under the Receiver Operating Characteristic curve} It measures how well a probe separates hallucinating generations from correct generations. A higher AUROC indicates better separation between hallucinated and correct generations. To evaluate the probe model, we use 5-fold stratified cross-validation with StratifiedKFold, shuffle, and fixed seed, scored by ROC-AUC. The reported AUROC is the mean across folds. 

In our experiments, we use four datasets. MedMCQA, our first dataset, is a medical multiple-choice Q\&A dataset \citep{pal2022medmcqa} designed to address real-world medical entrance exam questions. It has more than 194k multiple-choice questions and covers 2.4k healthcare topics and 21 medical subjects. The second dataset is MedQA-USMLE \citep{jin2021disease}, the English portion of MedQA collected from professional medical board exams with 12,723 questions. The third one is PubMedQA-labeled \citep{jin-etal-2019-pubmedqa}, a 1000 expert-labeled subset of PubMedQA. Finally, we use PubMedQA-artificial, the automatically generated subset of PubMedQA. It is much larger than PQA-L with 211.3k artificially generated instances. For all these datasets, we use the standard splits that are common in the literature for our experiments.

To work with these datasets, we selected four LLMs. BioMistral-7B\citep{labrak-etal-2024-biomistral}, which is a biomedical LLM built by further pre-training Mistral-7B-Instruct on PubMed and other sources. It is best viewed as a biomedical research model. Meditron-7B \citep{chen2024meditron} that is a medical model adapted from Llama-2-7B on data including PubMed and RedPajama-v1. Our third model is PMC-LLaMA-7B, a LLaMA-7B-based medical model adapted toward biomedical language. This model is fine-tuned on the PMC papers in the S2ORC dataset and focuses on medical-domain Q\&A. Finally, we used Llama-3.1-8B-Instruct. Unlike the other three, this one is a general-purpose instruction-tuned model, mainly designed for chat and similar applications \citep{grattafiori2024llama}.

We select these models because they are open source, widely used, and therefore suitable for both mechanistic analysis and community-relevant evaluation. Since our focus is the medical domain, we include three medical models. However, because medical models may encode domain-specific signals that affect hallucination detection, we also include Llama as a general-domain baseline to assess whether our findings transfer when the same medical questions are asked of a generic model. Our goal is not to identify hallucination signals that transfer between medical and general-domain models. Rather, this comparison is included for completeness and to make the results more reliable. It should be noted that our investigation is not limited to the setup reported here; we present a representative subset of results that captures the main patterns and omit additional configurations that do not change the overall message. Some of these additional results are in the appendix.

\section{Experimental Results}\label{exp}
Table \ref{tab:main_results} summarizes our results obtained on hallucination detection. Probe AUROC measures the performance of the probe on selected  neurons, while full-layer AUROC shows performance when all neurons of the layer are used. The random \(k=500\) baseline tests the impact of similarly sized random neuron subsets.
\begin{table}[htbp]
\centering
\begin{tabular}{lccccc}
\toprule
\shortstack{Model} &
\shortstack{Best\\layer} &
\shortstack{Probe\\AUROC} &
\shortstack{Full-layer\\AUROC} &
\shortstack{Random\\k=500 AUROC} &
\shortstack{Base hallu.\\rate} \\
\midrule
BioMistral-7B & L17 & 0.808 & 0.769 & 0.807 & 0.446 \\
Meditron-7B & L14 & 0.796 & 0.782 & 0.789 & 0.557 \\
PMC-LLaMA-7B & L10 & 0.767 & 0.739 & 0.763 & 0.573 \\
Llama-3.1-8B-Instruct & L15 & 0.863 & 0.815 & 0.859 & 0.208 \\
\bottomrule
\end{tabular}
\caption{\label{tab:main_results}
Hallucination-detection results for each LLM. Base hallu. rate is the percentage of the model’s original answers that were labeled as hallucinated before applying any detector.}
\end{table}

\begin{figure}[H]
\centering
\begin{subfigure}{0.245\textwidth}
\centering
\includegraphics[width=\linewidth]{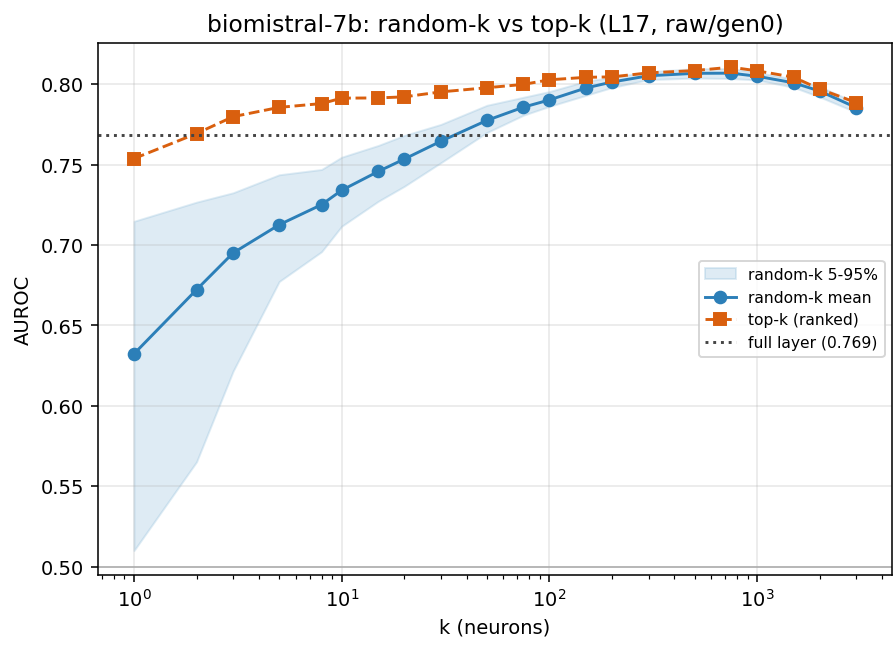}
\caption{Biomistral-7B}
\label{fig:image1}
\end{subfigure}
\hfill
\begin{subfigure}{0.245\textwidth}
\centering
\includegraphics[width=\linewidth]{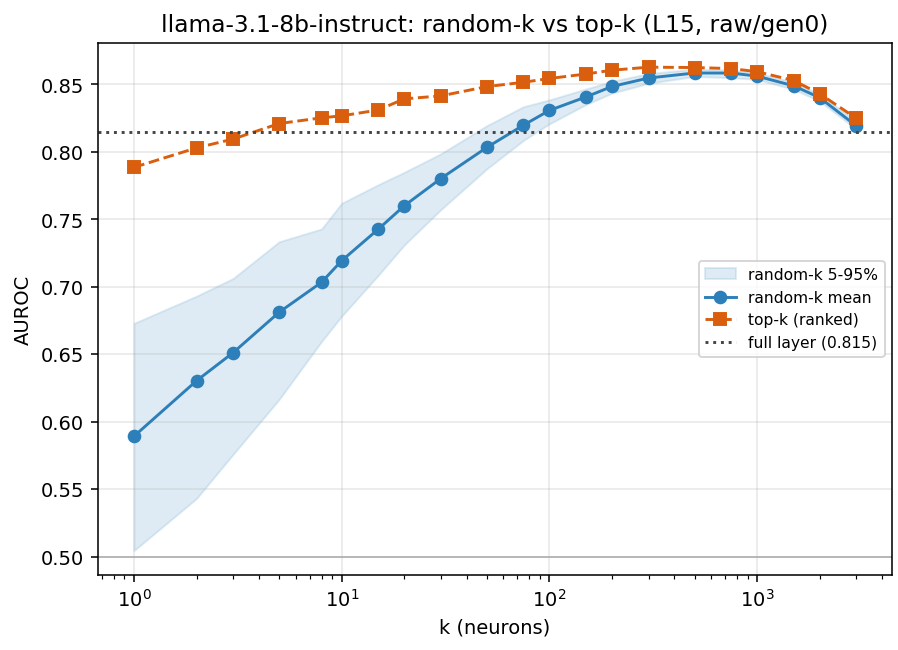}
\caption{Llama-3.1-8B}
\label{fig:image2}
\end{subfigure}
\hfill
\begin{subfigure}{0.245\textwidth}
\centering
\includegraphics[width=\linewidth]{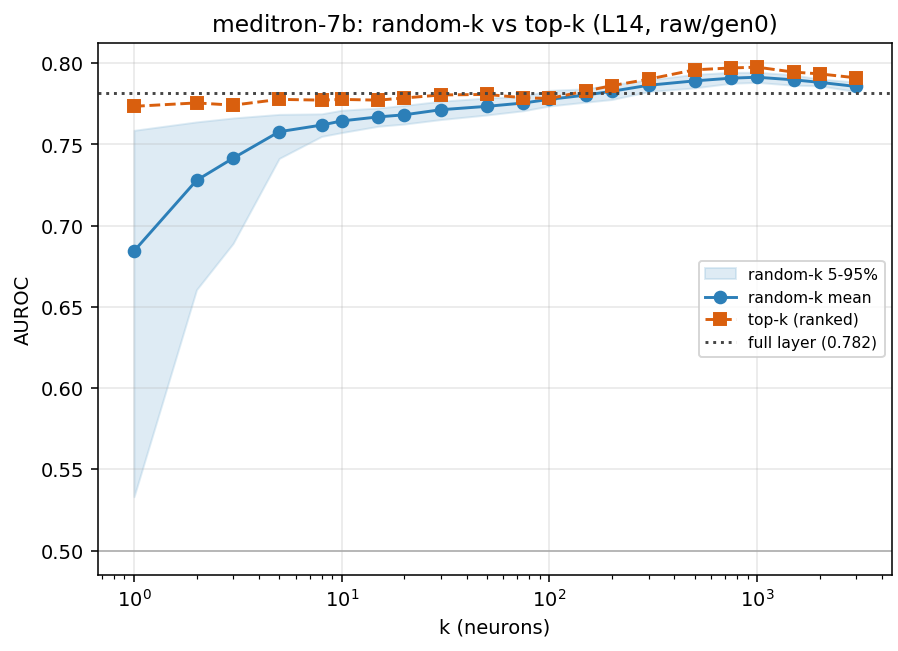}
\caption{Meditron-7B}
\label{fig:image3}
\end{subfigure}
\hfill
\begin{subfigure}{0.245\textwidth}
\centering
\includegraphics[width=\linewidth]{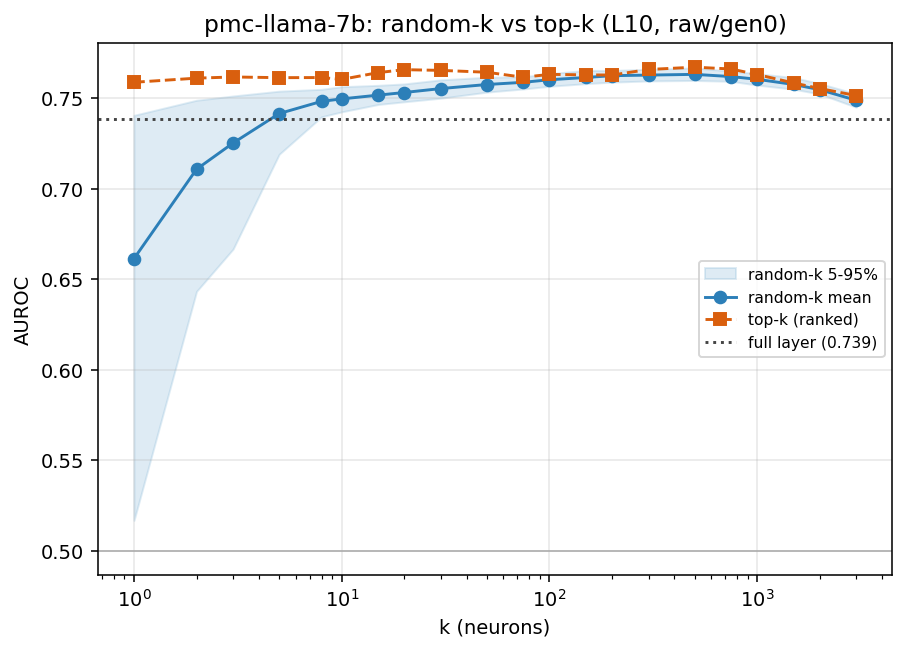}
\caption{PMC-LLaMA-7B}
\label{fig:image4}
\end{subfigure}
\caption{Detection AUROC as a function of $k$ (log scale) for each LLM's best $\texttt{Gen0}$ layer. The x-axis is $k$ and the y-axis is AUROC. Orange squares are top-$k$ neurons selected by individual discriminative AUROC; blue circles are the mean over random-$k$ subsets, with the shaded band the 5--95\% range across random draws; the dotted line is the full-layer probe. Selected neurons lead only at very small $k$; random subsets catch up by $k\sim300$, and both exceed the full layer before declining. A larger visualization is available in Appendix~\ref{app:ksweep_large}.}\label{fig:k_effect}
\end{figure}

Results show a consistent detection pattern across models. The best-layer probe AUROC ranges from 0.767 to 0.863, indicating that hallucination-related information is reliably detectable. However, using the entire layer is not always optimal. In every model, a random subset of (k=500) neurons matches or improves on the full-layer result. For example, the full-layer AUROC of BioMistral is 0.769, while the random subset reaches 0.807. These results also suggest that the signal is distributed rather than localized to a small set of neurons. Random subsets of a few hundred neurons are enough to recover nearly the best detection performance. Neuron selection matters mainly at small k values, especially below k $\sim$ 50, but selected and random subsets converge by around k $\approx$ 300. The random-k (=500) results are especially important because they show that adding more neurons does not always improve detection. Instead, the detection curve has an inverted-U shape. Performance peaks around k $\sim$ 300 to 750, then declines. See Figure \ref{fig:k_effect} for the visualization of this finding. 

To confirm our estimates are stable under resampling rather than driven by a single favorable split, we computed bootstrap 95\% confidence intervals for all models and conditions. The intervals are narrow across all models and conditions and are stable under resampling. For details see Appendix~\ref{app:bootstrap_ci}. In addition to our own model results, Table~\ref{tab:comp} places our probe in context by comparing it with several established hallucination-detection baselines. We include this comparison to show how our representation-based probe performs relative to common detection approaches.
\begin{table}[htbp]
\centering
\begin{tabular}{lcccc}
\toprule
Method & BioMistral & Meditron & PMC-LLaMA & Llama-3.1 \\
\midrule
\textbf{Our Method} & 0.808 & 0.796 & 0.767 & 0.863 \\
\midrule
\multicolumn{5}{l}{\textbf{Baseline models}} \\
\texttt{token\_neg\_prob} & 0.7358 & 0.4967 & 0.2663 & 0.7644 \\
\texttt{token\_entropy} & 0.6025 & 0.3068 & 0.2630 & 0.7061 \\
\texttt{saplma} \citep{azaria2023internal}& 0.7469 & 0.7729 & 0.7309 & 0.7916 \\
\texttt{haloscope} \citep{du2024haloscope} & 0.7243 & 0.7563 & 0.7453 & 0.6771 \\
\texttt{ccs} \citep{burns2022discovering} & 0.7154 & 0.7518 & 0.7376 & 0.6349 \\
\bottomrule
\end{tabular}
\caption{\label{tab:comp}
Comparison with baseline hallucination-detection methods. All scores in this table report probe AUROC. \texttt{token\_neg\_prob} uses token negative probability as a confidence-based signal. \texttt{token\_entropy} uses token entropy as an uncertainty-based signal.
}
\label{tab:baseline_comparison}
\end{table}
\vspace{-5mm}
\subsection{Cross-Position Circuit Structure}
\label{app:cross_position}
After establishing that hallucination is detectable from internal activations, and that this signal is not limited to a small fixed set of neurons, we next ask whether the underlying representation remains consistent across token positions. In other words, even if the most important neurons change from one position to another, the model may still preserve a similar hallucination-related activation geometry. This analysis therefore shifts from asking which neurons are selected to asking whether the broader representation is stable. We compare neuron-level overlap with representation-level similarity in Table~\ref{tab:cross_position_circuit_structure}.
\begin{table}[htbp]
\centering
\begin{tabular}{lcccc}
\toprule
Metric & BioMistral & Meditron & PMC-LLaMA & Llama-3.1 \\
\midrule
Jaccard top-50 & 0.0096 & 0.0320 & 0.0180 & 0.0041 \\
Random Jaccard & 0.0012 & 0.0038 & 0.0038 & 0.0012 \\
Spearman (AUROC rank) & 0.0520 & 0.0976 & 0.0651 & 0.0285 \\
CKA top-500 & 0.9514 & 0.8962 & 0.8620 & 0.9239 \\
\bottomrule
\end{tabular}
\caption{
Cross-position circuit-structure analysis. Jaccard top-50 measures overlap between the top 50 selected neurons across positions. Random Jaccard repeats the same process for random overlaps. Spearman measures whether neurons have similar AUROC-based importance rankings across positions. CKA top-500 measures whether the broader activation geometry is similar across selected neuron sets.
}
\label{tab:cross_position_circuit_structure}
\end{table}
\vspace{-1mm}

Results from the geometry analysis show that same top neurons rarely appear across positions. Although Jaccard top-50 is slightly above the random baseline, it stays very low for all models, and the Spearman correlations indicate that neurons important at one position are usually not important at another. Taken alone this might suggest an unstable/random signal, however, the high CKA top-500 values point to a different conclusion: the model preserves a similar overall representation while using different position-specific neurons. This suggests that the hallucination signal is not tied to fixed neuron identities. Instead, it appears to be encoded as a consistent representation-level pattern that can be expressed through different subsets of neurons at different token positions.

\subsection{Intervention Results}
\label{subsec:intervention_results}
As a final evaluation, we test whether the neurons that support hallucination detection can also be used for causal intervention. This asks whether the detected signal is only useful for reading out hallucination risk, or if directly intervening on the selected neurons can reliably change the model's answers.
\begin{table}[htbp]
\centering
\begin{tabular}{lcc}
\toprule
Quantity & Estimate & 95\% CI \\
\midrule
Selected-set swing & 0.012 & {[}0.007, 0.019{]} \\
Random-set swing & 0.008 & {[}0.005, 0.010{]} \\
$\Delta$ (selected-set $-$ random-set) & +0.0042 & {[}-0.0002, 0.0096{]}\\ \hline
Amplification effect & +0.0037 & {[}-0.0032, 0.012{]} \\
Logit slope, $P(\mathrm{correct})$ vs. $\alpha$ & -0.0045 & {[}-0.033, 0.020{]} \\
\bottomrule
\end{tabular}
\caption{
Intervention-effect summary. Selected-set swing measures how much the model's output changes when intervening on selected hallucination-associated neurons. Random-set swing measures the same effect for randomly chosen neurons. $\Delta$ estimates the additional effect of using selected neurons. Amplification and logit-slope analyses test whether increasing intervention strength produces a reliable change in correctness.
}
\label{tab:intervention_results}
\end{table}

As shown, the selected neurons do impact the model's output, but random neurons produce a comparable swing too, and the advantage of selected over random is relevantly small with a 95\% CI spanning zero ($\Delta = +0.0042$, $[-0.0002, 0.0096]$). One natural concern is that this null result could reflect interventions that are simply too weak to move the model. Our amplification analysis argues against this explanation. As the steering coefficient $\alpha$ increases, the correctness response remains flat (logit slope $-0.0045$, $[-0.033, 0.020]$). Therefore, within the tested range, making the intervention stronger does not convert the detectable hallucination signal into a reliable change in answer correctness.

We assessed significance using Holm-corrected McNemar tests. Each test compared targeted steering with control conditions on the same examples, with correction across models and datasets. After correction, none of the 16 model--dataset comparisons were significant. However, statistical power is not uniform across comparisons. Because the intervention uses a 20\% held-out split, low-base-rate datasets contain relatively few hallucinated examples in the held-out set. In several PubMedQA-artificial settings, this leaves fewer than 50 positive examples, which makes those comparisons underpowered by construction. For this reason, the null result is most reliable on the higher-base-rate datasets, where the tests are better powered and still show no advantage for targeted steering.

Across the intervention variants we tested, detection strength did not predict causal effect. This supports a separation between reading out hallucination risk and controlling model behavior. The signal is useful for detection, but it does not provide a reliable target for our single-layer neuron-level steering interventions. This does not rule out other forms of control, such as multi-layer, subspace-level, or circuit-level interventions. It only shows that high AUROC does not, by itself, imply controllability through direct neuron-level steering.

\section{Conclusion}
We investigated whether hallucination-related signals in medical language models are not only detectable, but also actionable through intervention. Across different settings, a simple linear probe consistently identified hallucination-related information, showing that hallucinations are readable from internal activations. However, this signal does not follow the sparse-neuron picture suggested by some prior work. Random neuron subsets often matched carefully selected subsets once a few hundred neurons were included, and cross-position analyses showed that similar information can appear in different neuron groups. Together, these results suggest that medical hallucination is encoded as a distributed, redundant, and low-dimensional representation rather than a small set of specialized neurons. Most importantly, detectability did not translate into controllability. Neurons useful for detection did not consistently outperform random neurons as intervention targets, and stronger interventions did not reliably improve correctness. Therefore, reading out a hallucination signal does not mean we can control it through standard neuron-level steering. Although hallucination-related information is clearly present in model activations, how these distributed representations drive model behavior remains an open question.

\section*{Limitations and Future Work}
This study has several limitations. Interventions were evaluated mainly at the layer with the strongest detection signal, and generation length was restricted, which could potentially limit generalization to longer medical reasoning or multi-step answers. The conclusions also apply only to the steering methods tested here. Future work should therefore explore interventions that better match the distributed structure of the signal, including multi-layer, subspace-level, and circuit-level methods, as well as combinations of internal representations with external grounding or fact-checking systems.

\section*{Ethics Statement}
This work studies hallucination in medical-domain LLMs, where fluent but incorrect outputs can create serious risks if treated as clinical advice. The models and methods evaluated here are intended only for research and analysis, not autonomous medical decision-making. We use established medical QA benchmarks rather than private patient data, but these datasets may still reflect biases from medical exams, biomedical literature, and clinical practice. Results may therefore not generalize to all populations, specialties, or real-world healthcare settings. Finally, our results show that strong hallucination detection does not necessarily imply effective control. A high-AUROC detector should not be treated as a safety guarantee. Medical language models still require human oversight and external verification/fact-checking in high-stakes settings.


\bibliography{colm2026_conference}
\bibliographystyle{colm2026_conference}
\newpage
\appendix
\section{Appendix}\label{appen}

\subsection{Data and Model Details}
\label{appen:data_model_info}
\begin{table}[htbp]
\centering
\small
\setlength{\tabcolsep}{5pt}
\begin{tabular}{lllll}
\toprule
Field & BioMistral-7B & Meditron-7B & PMC-LLaMA-7B & Llama-3.1-8B-Instruct \\
\midrule
Base & Mistral-7B & Llama-2-7B & Llama-7B & Llama-3.1-8B \\
Type & medical & medical & medical (papers) & general \\
$d_{\mathrm{inter}}$ & 14336 & 11008 & 11008 & 14336 \\
$N$ (clean) & 21008 & 20912 & 20717 & 21169 \\
Overall hall-rate & 0.446 & 0.557 & 0.573 & 0.208 \\
\bottomrule
\end{tabular}
\caption{
Model-level information for the four LLMs. $d_{\mathrm{inter}}$ indicates MLP intermediate dimension and $N$ is the number of clean evaluation examples.
}
\label{tab:app_model_info}
\end{table}

\begin{table}[htbp]
\centering
\small
\setlength{\tabcolsep}{6pt}
\renewcommand{\arraystretch}{1.1}
\begin{tabular}{llccc}
\toprule
Dataset & Model & $n$ & Hallu. rate & \#Pos. \\
\midrule
medmcqa & BioMistral-7B & 4944 & 0.614 & 3035 \\
medmcqa & Meditron-7B & 4995 & 0.711 & 3549 \\
medmcqa & PMC-LLaMA-7B & 5000 & 0.713 & 3564 \\
medmcqa & Llama-3.1-8B-Instruct & 5000 & 0.224 & 1120 \\
\midrule
medqa\_usmle & BioMistral-7B & 10174 & 0.574 & 5835 \\
medqa\_usmle & Meditron-7B & 10177 & 0.742 & 7548 \\
medqa\_usmle & PMC-LLaMA-7B & 10177 & 0.746 & 7596 \\
medqa\_usmle & Llama-3.1-8B-Instruct & 10177 & 0.294 & 2987 \\
\midrule
pubmedqa\_artificial & BioMistral-7B & 4918 & 0.045 & 221 \\
pubmedqa\_artificial & Meditron-7B & 4810 & 0.047 & 225 \\
pubmedqa\_artificial & PMC-LLaMA-7B & 4693 & 0.072 & 340 \\
pubmedqa\_artificial & Llama-3.1-8B-Instruct & 4992 & 0.018 & 89 \\
\midrule
pubmedqa\_labeled & BioMistral-7B & 972 & 0.291 & 283 \\
pubmedqa\_labeled & Meditron-7B & 930 & 0.349 & 325 \\
pubmedqa\_labeled & PMC-LLaMA-7B & 847 & 0.437 & 370 \\
pubmedqa\_labeled & Llama-3.1-8B-Instruct & 1000 & 0.207 & 207 \\
\bottomrule
\end{tabular}
\caption{
Dataset-level breakdown for each model. For each dataset--model pair, $n$ is the number of clean examples, Hallu. rate is the fraction of examples labeled as hallucinated, and \#Pos. is the number of positive hallucination labels.
}
\label{tab:app_dataset_breakdown}
\end{table}

\subsection{Cross-Dataset Detector Quality}
\label{app:cross_dataset_detector_quality}

This section evaluates how well each model's hallucination detector transfers across datasets. For each model, we train the detector on one dataset and test it on each of the four datasets. 

\subsubsection{BioMistral-7B}
\label{app:biomistral_cross_dataset}

\begin{table}[htbp]
\centering
\begin{tabular}{lcccc}
\toprule
Train / test & medmcqa & medqa\_usmle & pubmedqa\_art & pubmedqa\_lab \\
\midrule
medmcqa & 0.722 & 0.614 & 0.632 & 0.491 \\
medqa\_usmle & 0.647 & 0.661 & 0.774 & 0.551 \\
pubmedqa\_artificial & 0.499 & 0.474 & 0.954 & 0.737 \\
pubmedqa\_labeled & 0.539 & 0.494 & 0.764 & 0.868 \\
\bottomrule
\end{tabular}
\caption{
Cross-dataset hallucination-detection quality for BioMistral-7B. Rows indicate the training dataset and columns indicate the test dataset. Diagonal entries report same-dataset performance, Off-diagonal entries report transfer performance across datasets.
}
\label{tab:app_biomistral_cross_dataset}
\end{table}

\subsubsection{Meditron-7B}
\label{app:meditron_cross_dataset}

\begin{table}[htbp]
\centering
\begin{tabular}{lcccc}
\toprule
Train / test & medmcqa & medqa\_usmle & pubmedqa\_art & pubmedqa\_lab \\
\midrule
medmcqa & 0.612 & 0.511 & 0.696 & 0.502 \\
medqa\_usmle & 0.530 & 0.582 & 0.496 & 0.502 \\
pubmedqa\_artificial & 0.515 & 0.508 & 0.939 & 0.732 \\
pubmedqa\_labeled & 0.506 & 0.491 & 0.819 & 0.852 \\
\bottomrule
\end{tabular}
\caption{
Cross-dataset hallucination-detection quality for Meditron-7B. 
}
\label{tab:app_meditron_cross_dataset}
\end{table}

\subsubsection{PMC-LLaMA-7B}
\label{app:pmc_llama_cross_dataset}

\begin{table}[htbp]
\centering
\begin{tabular}{lcccc}
\toprule
Train / test & medmcqa & medqa\_usmle & pubmedqa\_art & pubmedqa\_lab \\
\midrule
medmcqa & 0.600 & 0.492 & 0.548 & 0.537 \\
medqa\_usmle & 0.466 & 0.586 & 0.413 & 0.452 \\
pubmedqa\_artificial & 0.510 & 0.505 & 0.778 & 0.564 \\
pubmedqa\_labeled & 0.503 & 0.495 & 0.568 & 0.719 \\
\bottomrule
\end{tabular}
\caption{
Cross-dataset hallucination-detection quality for PMC-LLaMA-7B.
}
\label{tab:app_pmc_llama_cross_dataset}
\end{table}

\subsubsection{Llama-3.1-8B-Instruct}
\label{app:llama_31_cross_dataset}

\begin{table}[htbp]
\centering
\begin{tabular}{lcccc}
\toprule
Train / test & medmcqa & medqa\_usmle & pubmedqa\_art & pubmedqa\_lab \\
\midrule
medmcqa & 0.859 & 0.762 & 0.856 & 0.630 \\
medqa\_usmle & 0.821 & 0.785 & 0.695 & 0.608 \\
pubmedqa\_artificial & 0.585 & 0.593 & 0.974 & 0.709 \\
pubmedqa\_labeled & 0.555 & 0.543 & 0.903 & 0.834 \\
\bottomrule
\end{tabular}
\caption{
Cross-dataset hallucination-detection quality for Llama-3.1-8B-Instruct.
}
\label{tab:app_llama_31_cross_dataset}
\end{table}

\subsection{Precision at Fixed Recall}
\label{app:precision_fixed_recall}
This section shows how precise the detector is when we force it to catch a fixed fraction of hallucinations. Results are reported in Table \ref{tab:precision_fixed_recall}. 
\begin{table}[H]
\centering
\begin{tabular}{lcccc}
\toprule
Recall & BioMistral & Meditron & PMC-LLaMA & Llama-3.1 \\
\midrule
0.50 & 0.7184 & 0.7634 & 0.7610 & 0.5832 \\
0.70 & 0.6840 & 0.7499 & 0.7480 & 0.5128 \\
0.80 & 0.6643 & 0.7465 & 0.7441 & 0.4710 \\
0.90 & 0.6369 & 0.7390 & 0.7384 & 0.4144 \\
\bottomrule
\end{tabular}
\caption{
Precision at fixed recall for the best $\texttt{Gen0}$ condition. Each model column reports precision at that recall level (the fraction of flagged examples that are actually hallucinations).
}
\label{tab:precision_fixed_recall}
\end{table}

This is useful for practical settings because increasing recall helps capture more hallucinations, however, it can also create more false alarms. Precision decreases as recall increases from 0.50 to 0.90, which is the expected trade-off. Meditron and PMC-LLaMA perform best and stay close to 0.74 precision even at 0.90 recall. BioMistral is slightly lower but still reasonably strong. Llama-3.1 has lower precision, especially at high recall, which means it produces more false positives when tuned to detect most hallucinations. Given these results, the detector can be useful in practice.

\subsection{Layer and Neuron Selection}
\label{app:layer_neuron_selection}
This section explains how we selected the strongest layers and probe settings for each model. We compare two activation views, $\texttt{raw}$ and $\texttt{exact\_cett}$, at two positions: a) the prompt position and b) the first generated token position ($\texttt{Gen0}$). For each condition, Table \ref{tab:app_neuron_probe_view_position} reports the best AUROC and the corresponding layer in parentheses. The final row gives the strongest AUROC found for each model across all listed conditions.

\begin{table}[htbp]
\centering
\small
\setlength{\tabcolsep}{5pt}
\begin{tabular}{lcccc}
\toprule
Probe & BioMistral & Meditron & PMC-LLaMA & Llama-3.1 \\
\midrule
$\texttt{raw}$ / prompt & 0.7904 (L24) & 0.7931 (L18) & 0.7676 (L3) & 0.8426 (L25) \\
$\texttt{raw}$ / $\texttt{Gen0}$ & 0.8084 (L17) & 0.7960 (L14) & 0.7671 (L10) & 0.8626 (L15) \\
exact\_cett / prompt & 0.7890 (L20) & 0.7960 (L17) & 0.7666 (L2) & 0.8426 (L28) \\
exact\_cett / $\texttt{Gen0}$ & 0.8069 (L17) & 0.7907 (L19) & 0.7684 (L8) & 0.8601 (L15) \\
\midrule
Best & 0.8084 & 0.7960 & 0.7684 & 0.8626 \\
\bottomrule
\end{tabular}
\caption{
Neuron-probe AUROC by activation view and position. The layer in parentheses is the best-performing layer for that condition.
}
\label{tab:app_neuron_probe_view_position}
\end{table}

Overall, the best performance is usually obtained from $\texttt{Gen0}$ activations, showing that the first generated token carries a strong hallucination-detection signal. The best layers are generally in the middle-to-later parts of the models, though the exact layer varies by architecture and activation view.

\subsection{Bootstrap 95\% Confidence Intervals}
\label{app:bootstrap_ci}
A single AUROC point estimate can look strong yet be unstable if small changes in the evaluation set shift performance. We therefore report bootstrap 95\% confidence intervals to quantify the uncertainty of each AUROC value.

\begin{table}[htbp]
\centering
\begin{tabular}{lcccc}
\toprule
Condition & BioMistral & Meditron & PMC-LLaMA & Llama-3.1 \\
\midrule
$\texttt{raw}$/prompt CI & {[}0.7845, 0.7958{]} & {[}0.7867, 0.7996{]} & {[}0.7603, 0.7739{]} & {[}0.8366, 0.8480{]} \\
$\texttt{raw}$/$\texttt{Gen0}$ CI & {[}0.8026, 0.8144{]} & {[}0.7896, 0.8028{]} & {[}0.7601, 0.7737{]} & {[}0.8568, 0.8678{]} \\
$\texttt{exact\_cett}$/prompt CI & {[}0.7835, 0.7950{]} & {[}0.7899, 0.8024{]} & {[}0.7589, 0.7725{]} & {[}0.8370, 0.8480{]} \\
$\texttt{exact\_cett}$/$\texttt{Gen0}$ CI & {[}0.8013, 0.8130{]} & {[}0.7843, 0.7971{]} & {[}0.7611, 0.7746{]} & {[}0.8550, 0.8651{]} \\
\bottomrule
\end{tabular}
\caption{
Bootstrap 95\% confidence intervals for AUROC under raw and exact\_cett evaluation settings. Intervals are reported for prompt and $\texttt{Gen0}$ conditions across all four models.
}
\label{tab:bootstrap_ci}
\end{table}
The intervals are narrow across all models and conditions, which indicate the AUROC estimates are stable under resampling. For BioMistral and Llama-3.1, the $\texttt{Gen0}$ intervals are consistently higher than the corresponding prompt intervals, so $\texttt{Gen0}$ representations provide a stronger detection signal; PMC-LLaMA shows very similar prompt and $\texttt{Gen0}$ intervals.

\subsection{Top-$k$ vs. Random-$k$ Neurons}
\label{app:topk_randomk_neurons}

We compare selected top neurons with randomly chosen neuron sets of the same size under the best $\texttt{Gen0}$ condition. The goal is to test whether the selected neurons provide a clear advantage over a random subset. We also report results from an L1 sparse probe, which encourages the detector to rely on fewer neurons. Table \ref{tab:app_topk_randomk_l1} summarizes the results for this analysis. 
\begin{table}[htbp]
\centering
\small
\setlength{\tabcolsep}{6pt}
\begin{tabular}{lcccc}
\toprule
Metric & BioMistral & Meditron & PMC-LLaMA & Llama-3.1 \\
\midrule
best AUROC & 0.8084 & 0.7960 & 0.7671 & 0.8626 \\
random-$k$ AUROC & 0.8069 & 0.7890 & 0.7633 & 0.8584 \\
lift (raw) & +0.0016 & +0.0069 & +0.0038 & +0.0042 \\
lift (exact\_cett) & +0.0075 & +0.0075 & +0.0087 & +0.0130 \\
L1 nonzero neurons & 3806 & 1619 & 1767 & 3046 \\
L1 AUROC & 0.8039 & 0.7997 & 0.7660 & 0.8543 \\
\bottomrule
\end{tabular}
\caption{
Top-$k$ versus random-$k$ neuron comparison. Best AUROC is the performance of the selected top neurons and random-$k$ AUROC is the performance of a random neuron set of the same size. Lift measures the gain of selected neurons over the random comparison. The L1 rows report the number of nonzero neurons used by the sparse probe and its corresponding AUROC.
}
\label{tab:app_topk_randomk_l1}
\end{table}

\subsection{$\texttt{raw}$ vs. $\texttt{exact\_cett}$ Neuron Overlap}
\label{app:raw_exact_cett_overlap}

We investigated if $\texttt{raw}$ and $\texttt{exact\_cett}$ activations select the same important neurons. This helps determine whether the two activation views identify the same neuron-level signal or reach similar detection performance through different neuron sets. We measure this using Jaccard overlap between the selected neuron lists at the best $\texttt{Gen0}$ layer. See Table \ref{tab:app_raw_exact_cett_overlap} for these results. 

\begin{table}[htbp]
\centering
\small
\setlength{\tabcolsep}{5pt}
\begin{tabular}{lcccc}
\toprule
Jaccard & BioMistral (L17) & Meditron (L19) & PMC-LLaMA (L8) & Llama-3.1 (L15) \\
\midrule
top-50 & 0.0101 & 0.0101 & 0.0204 & 0.0417 \\
top-100 & 0.0256 & 0.0526 & 0.0256 & 0.0526 \\
top-500 & 0.0471 & 0.0718 & 0.0684 & 0.0753 \\
\bottomrule
\end{tabular}
\caption{
Jaccard overlap between neurons selected from $\texttt{raw}$ and $\texttt{exact\_cett}$ activations. Larger values indicate that the two activation views select more similar neuron sets and smaller values indicate that they rely on different individual neurons.
}
\label{tab:app_raw_exact_cett_overlap}
\end{table}

\subsection{Full-Size k-Sweep Figures}
\label{app:ksweep_large}

Figure~\ref{fig:k_effect} in the main text is shown at a smaller size due to space limitations. Here, we provide an aggregated view of all curves, showing the relationship between $k$ and detection performance.

\begin{figure}[htbp]
\centering
\includegraphics[width=0.8\textwidth]{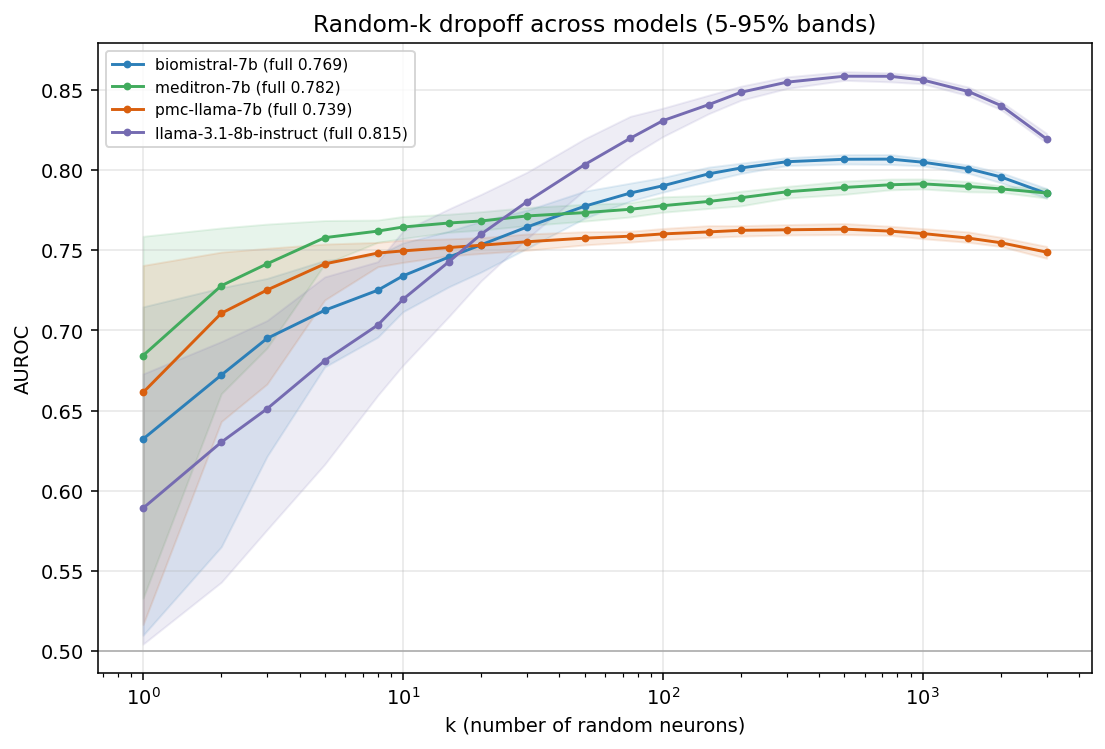}
\caption{Aggregated k-sweep results for all LLMs.}
\label{fig:ksweep_large_biomistral}
\end{figure}

\subsection{Position Fusion}
\label{app:position_fusion}
We studied whether hallucination detection works best from the prompt position, the first generated token position, or both together. We compare prompt-only features (P), $\texttt{Gen0}$-only features (G), and their concatenation (C). If concatenation improves performance, then the two positions provide complementary information and if not, one position already captures most of the useful signal. Results are reported in Table \ref{tab:app_position_fusion}.

\begin{table}[htbp]
\centering
\small
\resizebox{\textwidth}{!}{%
\begin{tabular}{lcccc}
\toprule
Dataset & BioMistral P / G / C & Meditron P / G / C & PMC-LLaMA P / G / C & Llama-3.1 P / G / C \\
\midrule
all & 0.780 / 0.784 / 0.793 & 0.772 / 0.777 / 0.779 & 0.766 / 0.763 / 0.764 & 0.834 / 0.833 / 0.855 \\
medmcqa & 0.680 / 0.707 / 0.706 & 0.598 / 0.576 / 0.591 & 0.571 / 0.553 / 0.559 & 0.828 / 0.851 / 0.865 \\
medqa\_usmle & 0.635 / 0.650 / 0.658 & 0.566 / 0.558 / 0.574 & 0.545 / 0.563 / 0.567 & 0.774 / 0.779 / 0.799 \\
pubmedqa\_artificial & 0.930 / 0.937 / 0.943 & 0.930 / 0.920 / 0.939 & 0.721 / 0.746 / 0.762 & 0.953 / 0.951 / 0.957 \\
pubmedqa\_labeled & 0.768 / 0.819 / 0.811 & 0.784 / 0.804 / 0.823 & 0.657 / 0.632 / 0.689 & 0.750 / 0.771 / 0.784 \\
\bottomrule
\end{tabular}%
}
\caption{
Position-fusion results by dataset and model.
}
\label{tab:app_position_fusion}
\end{table}

\subsection{Dimensionality and Calibration of the Hallucination Signal}
\label{app:dimensionality_calibration}

This section examines the shape and reliability of the learned hallucination signal. We ask whether the signal is concentrated in a small number of directions, whether nonlinear models add much beyond a simple linear summary, and whether the detector's probability scores are well calibrated.

\begin{table}[htbp]
\centering
\small
\setlength{\tabcolsep}{6pt}
\begin{tabular}{lcccc}
\toprule
Metric & BioMistral & Meditron & PMC-LLaMA & Llama-3.1 \\
\midrule
PC1 variance & 0.630 & 0.738 & 0.771 & 0.482 \\
PC1-only AUROC & 0.7725 & 0.7744 & 0.7625 & 0.7908 \\
non-linearity premium & +0.0344 & +0.0163 & +0.0059 & +0.0692 \\
Cohen's d (best neuron) & 0.9477 & 1.2800 & 1.1329 & 0.9241 \\
diff/correct norm ratio & 0.7039 & 0.8331 & 0.8112 & 0.3872 \\
Brier & 0.1739 & 0.1609 & 0.1707 & 0.1190 \\
ECE & 0.0304 & 0.0400 & 0.0200 & 0.0266 \\
\bottomrule
\end{tabular}
\caption{
Dimensionality, effect-size, and calibration statistics for the hallucination detector. PC1 variance and PC1-only AUROC measure how much of the signal is captured by one main direction in activation space. The non-linearity premium measures the gain from a more flexible detector over a simpler linear summary. Cohen's $d$ measures separation between hallucinated and correct examples for the best neuron. The diff/correct norm ratio compares the hallucination-related change to the correct-output reference. Brier and ECE measure calibration, where lower values indicate better probability calibration.
}
\label{tab:app_dimensionality_calibration}
\end{table}

\end{document}